\documentclass[10pt,twocolumn,letterpaper]{article}

\usepackage{cvpr}
\usepackage{times}
\usepackage{epsfig}
\usepackage{graphicx}
\usepackage{amsmath}
\usepackage{amssymb}
\usepackage{multirow}
\usepackage{makecell}
\usepackage{subcaption}
\usepackage{sidecap}
\sidecaptionvpos{figure}{t}
\usepackage[table]{xcolor}
\usepackage[group-separator={,}]{siunitx}
\usepackage{colortbl,hhline}
\usepackage{enumitem}
\usepackage{stmaryrd}
\usepackage{mathrsfs}
\usepackage[ruled,vlined]{algorithm2e}
\usepackage{slashbox}
\usepackage{bm}

\usepackage{amsthm}
 
\theoremstyle{definition}
\newtheorem{definition}{Definition}[section]

\def\tC{{{C}}} % total classes
\def\sC{{{S}}} % seen classes
\def\nC{{{T}}} % novel classes
\def\uC{{{U}}} % unseen classes
\def\fC{{{Q}}} % few-shot classes

\usepackage[pagebackref=true,breaklinks=true,letterpaper=true,colorlinks,bookmarks=false]{hyperref}

\cvprfinalcopy % *** Uncomment this line for the final submission

\def\cvprPaperID{5485} % *** Enter the CVPR Paper ID here

\ifcvprfinal\fi
\begin{document}

\title{Any-Shot Object Detection} % Replace with your title

% \author{Shafin Rahman$^{*\dagger}$, Salman Khan$^{\ddagger}$, Nick Barnes$^{*}$, Fahad Shahbaz Khan$^{\ddagger}$\\
% $^{*}$Australian National University, $^{\dagger}$North South University, \\
% $^{\ddagger}$Inception Institute of Artificial Intelligence\\
% {\tt\small {\{shafin.rahman,nick.barnes\}@anu.edu.au, \{salman.khan,fahad.khan\}@inceptioniai.org}}
% }

\author{Shafin Rahman$^{*\dagger\mathsection}$, Salman Khan$^{\ddagger}$, Nick Barnes$^{*}$, Fahad Shahbaz Khan$^{\ddagger}$\\
$^{*}$Australian National University, $^{\dagger}$North South University, $^{\mathsection}$Data61-CSIRO\\
$^{\ddagger}$Inception Institute of Artificial Intelligence\\
{\tt\small {\{shafin.rahman,nick.barnes\}@anu.edu.au, \{salman.khan,fahad.khan\}@inceptioniai.org}}
}

\maketitle

%%%%%%%%% ABSTRACT
\begin{abstract}
Previous work on novel object detection considers zero or few-shot settings where none or few examples of each category are available for training. In real world scenarios, it is less practical to expect that `\emph{all}' the novel classes are either unseen or {have} few-examples. Here, we propose a more realistic setting termed `\emph{Any-shot detection}', where totally unseen and few-shot categories can simultaneously co-occur during inference. Any-shot detection offers unique challenges compared to conventional novel object detection such as, a high imbalance between unseen, few-shot and seen object classes, susceptibility to forget base-training while learning novel classes and distinguishing novel classes from the background. To address these challenges, we propose a unified any-shot detection model, that can concurrently learn to detect both zero-shot and few-shot object classes. Our core idea is to use class semantics as prototypes for object detection, a formulation that naturally minimizes knowledge forgetting and mitigates the class-imbalance in the label space. Besides, we propose a rebalanced loss function that emphasizes difficult few-shot cases but avoids overfitting on the novel classes to allow detection of totally unseen classes. Without bells and whistles, our framework can also be used solely for Zero-shot detection and Few-shot detection tasks. We report extensive experiments on Pascal VOC and MS-COCO datasets where our approach is shown to provide significant improvements.
\end{abstract}

\section{Introduction}\vspace{-0.2em}
Traditional object detectors are designed to detect the categories on which they were originally trained. In several applications, such as self-driving cars, it is important to extend the base object detector with novel categories that were never seen before. The current `novel' object detection models proposed in the literature target either of the two distinct settings, \emph{Zero-shot detection} (ZSD) and \emph{Few-shot detection} (FSD). In the former setting, it is assumed that totally unseen objects appear during inference and a model must learn to adapt for novel categories using only their class description (semantics). In the latter setting, a small and fixed-number of novel class samples are available for model adaptation. However, in a practical scenario, restricting the novel classes to be always unseen (with zero visual examples) or always with few-shot examples can limit the generality of the model.

\begin{figure}[!tp]
    \centering
    \includegraphics[width=1\columnwidth,trim={0cm 0cm 2.7cm .5cm},clip]{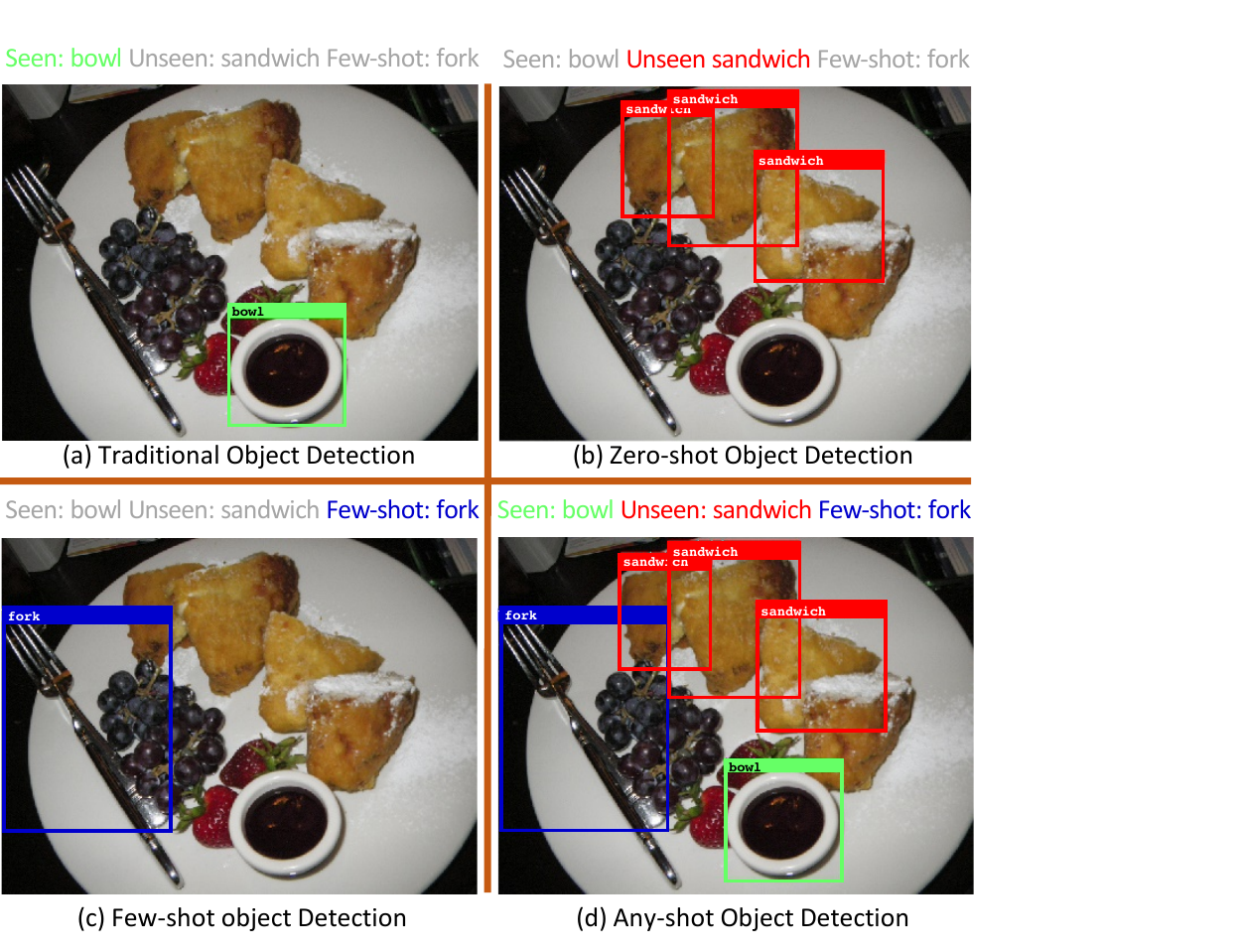}\vspace{-0.5em}
    % left lower right up
    \caption{\small (a) A traditional object detection method only detects seen objects. In the same vein, zero and few-shot object detection method{s} can detect (b) unseen or (c) few-shot objects. (d) Our proposed Any-shot detection method can \emph{simultaneously} detect seen, unseen and few-shot objects. }
    \label{fig:MainIdea}
\end{figure}

In a real-world scenario, both unseen and few-shot classes can be simultaneously of interest.  Moreover, we may encounter a few examples of a novel class that was previously supposed to be unseen. In such a case, an adaptive model must leverage from new information to improve its performance in an online fashion. To address these requirements, we introduce a new `\emph{Any-shot Detection}' (ASD) protocol where a novel class can have zero or a few training examples.  Since, the existing object detection models can either work for zero-shot or few-shot settings, we develop a unified framework to address the ASD problem (see Fig.~\ref{fig:MainIdea}). Remarkably, since the ASD task sits at the continuum between ZSD and FSD, our model can be directly applied to both these problems as well.

The ASD task poses new challenges for novel object detection. First, a high data imbalance between unseen, few-shot and seen classes can lead to a biased detection model. Additionally, the fine-tuning performed on few-shot examples can lead to forgetting previously acquired knowledge, thereby deteriorating model performance on seen and unseen classes. To overcome these challenges, we propose to learn a mapping from the visual space to the semantic space where class descriptors serve as fixed prototypes. The semantic prototypes in our approach encode inherent class relationships (thus enabling knowledge transfer from the seen to the unseen), helps us disentangle totally unseen concepts  from the background and can be automatically updated to align well with the visual information. Besides, we introduce a novel rebalancing loss function for the fine-tuning stage that functions on few-shot examples. This loss serves two objectives, i.e., to focus on the errors made for few-shot classes and at the same time avoid overfitting them so that it remains generalizable to totally unseen categories.

Our main contributions are:\vspace{-0.5em}
\begin{itemize}\setlength{\itemsep}{-0.2em}
\item A unified framework that can accommodate ZSD, FSD, ASD and their generalized settings.

\item Learning with semantic class-prototypes that are well aligned with visual information and help minimize forgetting old concepts.

\item An end-to-end solution with a novel loss function that rebalances errors to penalize difficult cases yet remains generalizable to unseen objects. 

\item Extensive experiments with new ASD setup, as well as comparisons with traditional FSD and ZSD frameworks demonstrating significant improvements.
\end{itemize}

%--------------------------------------------------------
\section{Related Work}
\textbf{N-shot recognition:} There exist three categories of methods for n-shot recognition. The \textit{first} body of work targets only zero-shot recognition (ZSR) tasks \cite{Xian_PAMI_2018}. They perform training with seen data and test on unseen (or unseen+seen) data. To relate seen and unseen classes, they use semantic embeddings e.g., attributes \cite{aPY_2009} or word vectors \cite{Mikolov_NIPS_2013,Jeffrey_Glove_2014}.
%(like word2vec \cite{Mikolov_NIPS_2013} or GloVe \cite{Jeffrey_Glove_2014}). 
{The} ZSR task has been investigated under popular themes such as transduction \cite{Song_2018_CVPR,Zhao_NIPS_2018}, domain adaptation \cite{Kodirov_2015_ICCV,Zhao_NIPS_2018}, adversarial learning \cite{Xian_2018_CVPR} and class-attribute association \cite{Al-Halah_2016_CVPR,al2017automatic}. 
The \textit{second} body of work targets only few-shot recognition (FSR) task \cite{Chen_ICLR_2019}. This task leverages few labeled examples to classify novel classes. Most popular methods to solve FSR are based on meta-learning where approaches perform metric learning to measure the similarity between input and novel classes \cite{Vinyals_NIPS_2016,Snell_NIPS_2017,Ravi_ICLR_2017}, adapt the meta-learner by calculating gradient updates for novel classes \cite{Ravi_ICLR_2017} or predict the classifier weights for novel classes \cite{Qi_2018_CVPR}. 
%Some other FSR methods attempts to augment data for the novel classes by generating hallucinated images \cite{Hariharan_2017_ICCV} and maps FSR as a domain adaptation problem to match the distribution of source (seen) and target (novel) classes \cite{Luo_NIPS_2017}.  %SK reducing paper length
The \textit{third} body of work addresses both zero and few-shot learning tasks together \cite{Schonfeld_2019_CVPR,Xian_2019_CVPR,Rahman_TIP_2018,ReViSE_CoRR_2017}. These approaches are the extended version of ZSR or FSR methods that consider word vectors to accommodate both problems within a single framework. Our current work belongs to the third category, but instead of a recognition task, we focus on the detection problem, that is significantly more challenging.

\textbf{Zero-shot detection:} Different from traditional object detection (where only seen objects are detected), ZSD aims to detect both seen and/or unseen objects. Pioneer{ing} works on  ZSD attempt to extend establish{ed} object detection methods to enable ZSD. For example,   \cite{Bansal_2018_ECCV}, \cite{Demirel_BMVC_2018,zhu2018zero} and \cite{rahman2018ZSD} employ pre-computed object proposals \cite{Edge_Boxes_2014}, YOLOv2 \cite{Redmon_yolo9000_2016} and Faster-RCNN \cite{Faster_RCNN_2017} based methods for ZSD, respectively. Recent methods for ZSD employ specialized polarity loss \cite{rahman2018polarity}, explore transductive settings \cite{Rahman_2019_ICCV} and use raw textual description instead of only class-names 
%word vectors 
\cite{Li_AAAI_2019}.
All the above methods focus on only ZSD and Generalized ZSD tasks but cannot accommodate FSD scenario when new instances of unseen images become available. In this paper, we propose a method that can perform ZSD, FSD, and ASD tasks seamlessly, including their generalized cases.

\begin{figure}[!t]
    \centering
    \includegraphics[width=1\columnwidth]{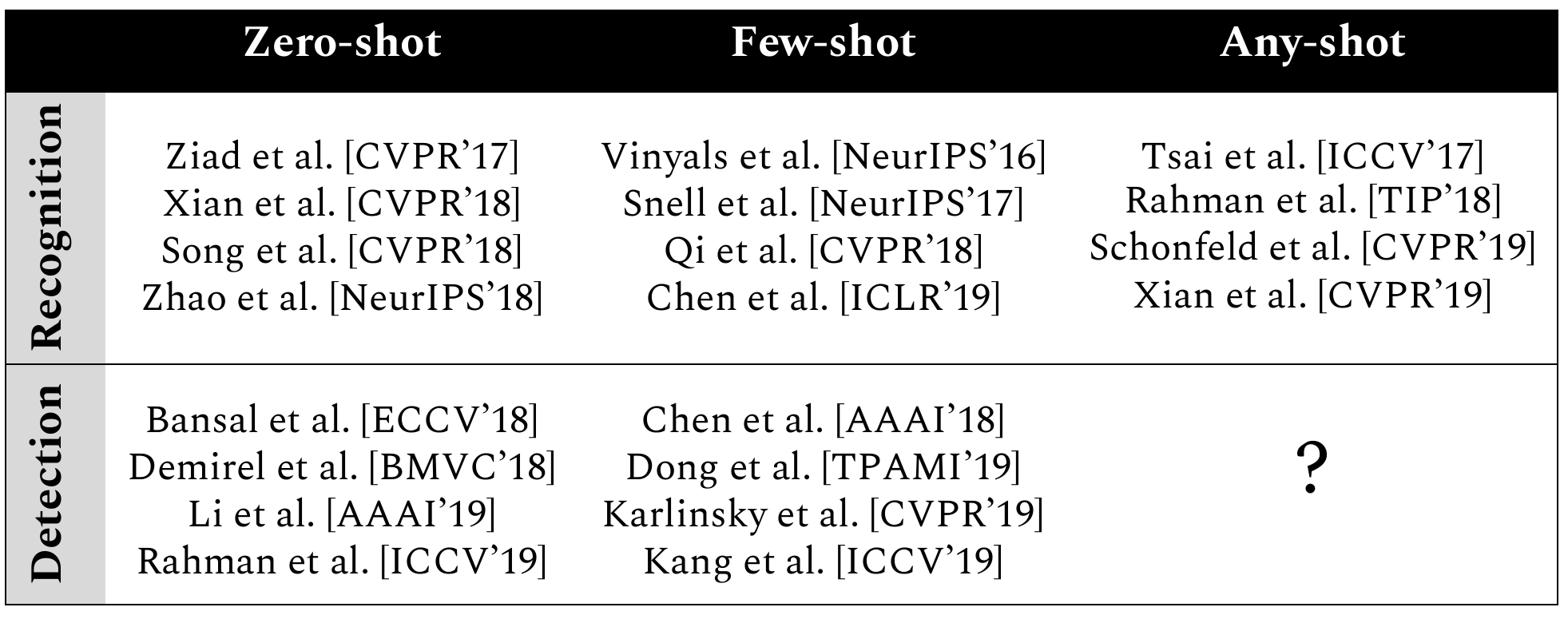}
    \vspace{-1em}
    \caption{{The} any-shot object detection (ASD) problem has not been addressed in the literature before. Importantly, an ASD system can automatically perform both ZSD and FSD, which no prior approach can simultaneously offer.  }
    \label{fig:rel_work}
\end{figure}

\textbf{Few-shot detection:} 
FSD methods attempt to detect novel classes for which only a few instances (1-10) are available during the inference stage \cite{Dong_PAMI_2019,Wang_2015_CVPR}. Among the early attempts of FSD, \cite{Chen_AAAI_2018} proposed a regularizer that works on standard object detection models to detect novel classes. Later, \cite{Karlinsky_2019_CVPR} proposed a distant metric learning-based approach that learn{ed}
representative vectors to facilitate FSD. 
The drawback of the above FSD methods is that they cannot handle seen/base classes during test time. Recently, \cite{Kang_2019_ICCV} proposed to train a base network with seen data and then fine-tune it by meta-network learning that predicts scores for both seen and novel classes by re-weighing the base network features. This approach can perform generalized FSD but cannot accommodate ZSD or ASD scenario. In this paper, we address the mentioned gap in the literature.

%-------------------------------------------------------------------
\section{Novel Object Detection}

Novel object detection refers to enhancing the ability of a traditional object detector model to detect a new set of classes that were not present during training. We propose a unified Any-shot Detection (ASD) setting\footnote{Our any-shot detection setting is different from \cite{Xian_2019_CVPR}, which considers zero and few-shot problems \emph{separately} 
for a simpler \emph{classification} task.} where novel classes include both few-shot and unseen (zero-shot) classes. This is in contrast to the existing works on novel object detection that treat zero and few-shot detection in an isolated manner. In the absence of the unseen class and few-shot classes, our problem becomes identical to a conventional FSD and ZSD problem, respectively. In this way, our proposed ASD settings unifies ZSD and FSD in a \emph{single} framework.

\subsection{Problem Formulation}
Assume a total of $\tC$ object classes are present in a given test set that need to be detected. Out of these,  $\sC (> 0)$ seen classes have many, $\fC (\geq 0)$ few-shot classes have few and $\uC (\geq 0)$ unseen classes has no examples available in the training dataset and $\tC = \sC{+}\fC{+}\uC$. Here, $\nC = \fC{+}\uC$ represents the total number of novel classes that become available during the inference stage. For each class, a semantic description is available as a $d$-dimensional embedding vector. The semantic embeddings of all classes are denoted by $\bm{W} = [\bm{W}_s, \bm{W}_f, \bm{W}_u] \in \mathbb{R}^{d\times \tC}$, where  $\bm{W}_s \in \mathbb{R}^{d \times \sC}$, $\bm{W}_f \in \mathbb{R}^{d \times \fC}$ and $\bm{W}_u \in \mathbb{R}^{d \times \uC}$ are semantic vectors for seen, few-shot and unseen classes, respectively. 

\begin{figure}[tp]
    \centering
    \includegraphics{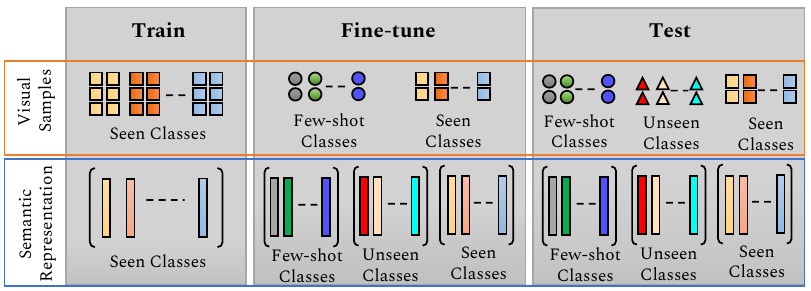}\vspace{-0.5em}
    \caption{\small An overview of Any-shot Detection setting.}
    \label{fig:setting}
\end{figure}

The base training set ($\mathcal{D}_{tr}$) contains $N_{tr}$ images with instances from $\sC$ seen classes. Each training image $\bm{I}$ is provided with a set of bounding boxes, where each box $\textbf{b}_{tr}$ is provided with a seen label $\textbf{y}_{tr} \in \{0,1\}^S$. Similarly, when $\fC {>} 0$, we have a fine-tuning dataset ($\mathcal{D}_{ft}$) with $N_{ft}$ images containing instances from both seen and few-shot classes for which bounding box $\textbf{b}_{ft}$ and class label $\textbf{y}_{ft} \in \{0,1\}^C$ annotations are present. During inference, we have a testing dataset $\mathcal{D}_{ts}$ with $N_{ts}$ images where each image can contain  any number of seen and novel class objects (see Fig.~\ref{fig:setting}).

Our task is to perform any-shot detection, defined as:
\vspace{-0.5em}
\theoremstyle{definition}
\begin{definition}{\emph{Any-shot detection:}}
 When $\fC {>} 0$ and $\uC {>} 0$, predict object labels and associated bounding boxes for $\nC$ novel classes, that include both zero and few-shot classes.
\end{definition}\vspace{-0.5em}
In comparison, the traditional zero and few-shot problems can be defined as: (a) \textbf{Few-shot detection:} When $\fC {>} 0$ but $\uC {=} 0$, predict object labels and associated bounding boxes for all $\fC$ classes. (b) \textbf{Zero-shot detection:} When $\fC {=} 0$ but $\uC {>} 0$, predict object labels and associated boxes for all $\uC$ classes. Note, if $\fC {=} \uC {=} 0$ then the problem becomes equivalent to a traditional detection task.

We also study the generalized ASD problem defines as: \vspace{-0.5em}
\theoremstyle{definition}
\begin{definition}{\emph{Generalized any-shot detection:}}
 When $\{\sC, \fC, \uC \}{\subset} \mathbb{Z}^{+}$, 
 %$\sC {>} 0$, $\fC {>} 0$ and $\uC {>} 0$, 
 predict object labels and box locations for all $\tC$ classes, that include seen, zero and few-shot classes.
\end{definition}\vspace{-0.5em}
In the same vein, generalized zero and few-shot detection problems aim to detect seen classes in addition to novel ones. Next, we describe our approach for ASD and GASD.

\subsection{Method}
Our goal is to design a model that can \emph{simultaneously} perform zero, few, and many-shot detection. %During training, t
This model is trained with seen classes, and is quickly adapted during inference for zero-shot and few-shot classes. 
This problem set-up has the following major challenges: \emph{(a) Adaptability:} A trained model must be flexible enough to incorporate new classes (with no or few examples) on the go, \emph{(b) Learning without Forgetting:} While the model is adapted for new classes, it must not forget the previous knowledge acquired on seen classes, and \emph{(c) Class Imbalance:} The classes representations are highly imbalanced: some with many instances, some with none and others with only a few. Therefore, the learning regime must be robust against the inherent imbalance in {the} ASD setting.

 {At}
a high-level, our proposed approach has two main components that address the above mentioned problems. First, we consider the semantic class prototypes to serve as anchors in the prediction space, thereby providing the flexibility to extend to any number of novel classes without the need to retrain network parameters. We show that such a representation also helps in avoiding catastrophic forgetting that is likely to occur otherwise. Furthermore, we propose a new loss formulation to address the class imbalance problem, that specifically focuses on difficult cases and minimizes model's bias against rare classes. We elaborate the novel aspects of our approach below.

\begin{figure*}[!t]
    \centering
    \includegraphics[width=1\textwidth,trim={0cm 0cm 0cm 0cm},clip]{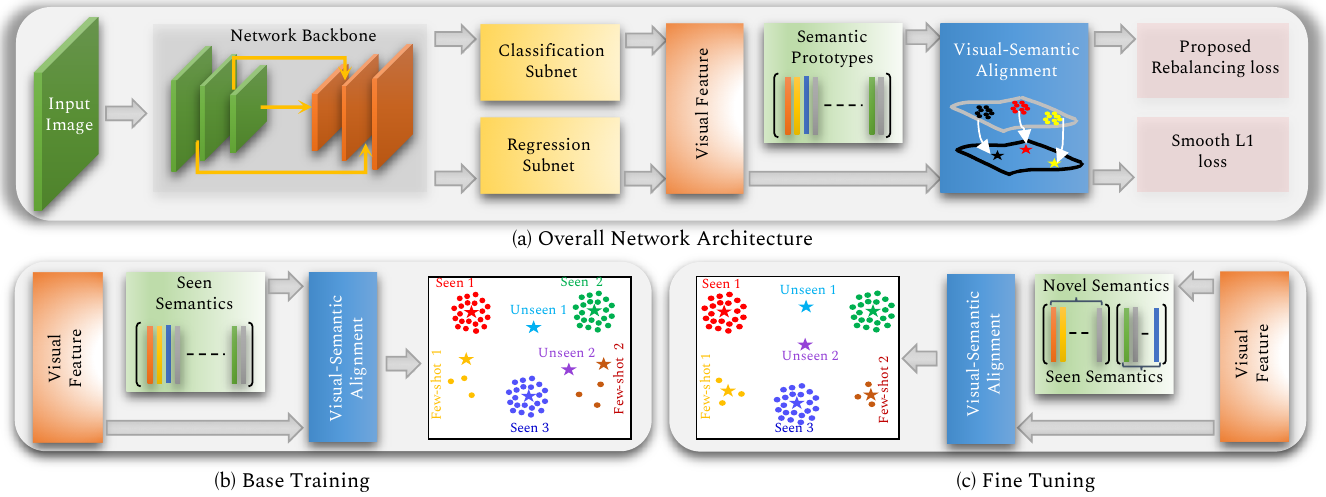}\vspace{-1em}
    % left lower right up
\caption{\small (a) Network architecture. The visual-semantic alignment is performed using (b) seen semantic{s} during base training and with (c) seen and novel semantics during fine-tuning. The visual features from {the} classification and regression units are separately used for visual-semantic alignment and subsequent loss calculation. 
}\vspace{-0.7em}
    \label{fig:arch}
\end{figure*}

\subsubsection{Learning without Forgetting}\label{sec:dontforget}
Traditional object detection models are static approaches that cannot dynamically adapt to novel classes. The flexibility to introduce novel object classes, after the base model training, requires special consideration, e.g., by posing it as an incremental learning problem \cite{li2018learning,Chen_NIPS_2017,shmelkov2017incremental}. In such cases, since classifier weights for novel categories are learned from scratch, the knowledge distillation concept \cite{hinton2015distilling} is applied to avoid forgetting the old learning. 
Such a strategy is not useful in our case because unlike previous approaches, we do not have access to many examples of the new task and a subset of novel classes has no training examples available.

To allow adding novel classes without forgetting old concepts, our approach seeks to disentangle the feature learning and classification stages. Precisely, we develop a training mechanism in which adding new classes does not require re-training base-network's parameters. This is achieved by defining the output decision space in the form of semantic class prototypes. These semantic class representatives are obtained in an unsupervised manner using a large text corpus, such as the Wikipedia, and encode class-specific attributes as well as the inter-class relationships \cite{Mikolov_NIPS_2013,Jeffrey_Glove_2014}. 

During base-model training, the model learns to map visual features to semantic space. At the same time, the semantic space is well-aligned with the visual concepts using a learnable projection. Note that only seen class semantics are used during the training stage. Once the base-model is trained, novel classes (both zero and few-shot) can be accommodated at inference time taking advantage of the semantic class descriptions of the new classes. For novel classes whose new few-examples are available during inference, we fine-tune the model to adapt semantic space, but keeping the base-model's architecture unchanged. Essentially, 
this means that adding new classes does not demand any changes to the architecture of {the} base-network. Still, the model is capable of generating predictions for novel classes since it has learned to relate visual information with semantic space during training (see Fig.~\ref{fig:arch}).

\begin{figure*}[!t]
  \centering.
   \includegraphics[width=.95\textwidth,trim={3cm 3.42cm 2.5cm 3cm},clip]{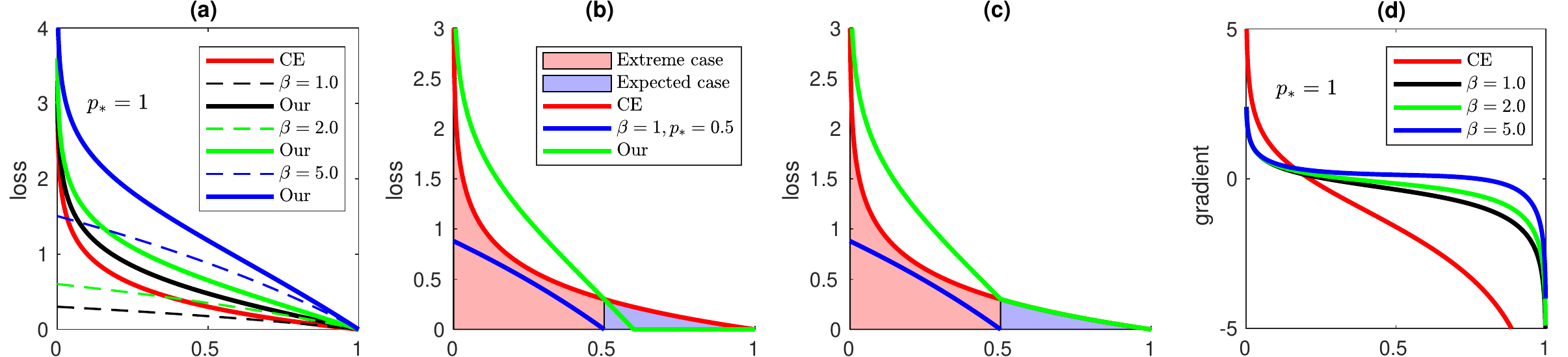}
   % left lower right up
   \vspace{-0.6em}
   \caption{\small Loss visualization. The colored dotted lines represent $\beta$-controlled penalty function, $h(.)$ and the solid lines represent loss functions. \textbf{(a)} The red line represents  {cross entropy} (CE) which is compared to our loss with $\beta = 1,2$ and $5$ shown as black, green and blue lines, respectively. \textbf{(b)} Our loss {(green line)} with fixed $p_* {=} 0.5$. Here, our loss can be less than CE {(red line)} for the expected case. \textbf{(c)} Our loss curve (green line) with dynamic $p_*$. Here, our loss calculates the same value as CE {(red line)} for the expected case. The red shaded region represents extreme case since $p {<} p_*$ and blue shaded region represents expected or moderate case $p {\geq} p_*$. (d) Derivatives of CE and our loss for different $\beta$. {See the appendix for the gradient analysis.}
   }
   \vspace{-0.7em}
\label{fig:loss_shape}
\end{figure*}

Formally, a training image $\bm{X}$ is fed to a deep network $f(\cdot)$, that produces a visual feature vector  $f(\bm{X}) \in \mathbb{R}^{n}$ corresponding to a particular box location. In a parallel stream, seen word vectors $\bm{W}_s$, are passed through a light-weight subnet denoted by $g(\cdot)$, producing the transformed word vectors $g(\bm{W}_s)$. Then, we connect the visual and semantic streams via a trainable layer, $ \bm{U} \in \mathbb{R}^{n \times d}$. Finally, a set of seen scores $\bm{p}_s$ is obtained by applying  {a} sigmoid activation function ($\sigma$). The overall computation is given by,
\begin{equation}\label{eq:pred_seen}
    \bm{p}_s = \sigma \big(f(\bm{X})^T \bm{U}  g(\bm{W}_s) \big).
\end{equation}
The mapping layer $\bm{U}$ can be understood as the bridge between semantic and visual domains. Given the visual and semantic mappings,  $f(\bm{X})$ and $g(\bm{W}_s)$ respectively, $\bm{U}$ seeks to maximize the alignment between the visual feature and its corresponding semantic class such that the prediction $\bm{p}_s$ is maximized. In a way, $\bm{p}_s$ is the alignment compatibility scores where a higher score means more compatibility between feature and semantics. The function $f(\cdot)$ can be implemented with a convolutional network backbone (e.g. ResNet \cite{ResNet_CVPR_2016}) and $g(\cdot)$ can be implemented as a fixed or trainable layer. In  {the} fixed case, fixed word vectors can be directly used for alignment i.e., $g(\bm{W}_{s}) = \bm{W}_{s}$. In the trainable case, $\bm{W}_s$ can be updated using a trainable metric $\bm{M} \in \mathbb{R}^{d \times v}$ and a word vocabulary  $\bm{D} \in \mathbb{R}^{v \times d}$, resulting in $g(\bm{W}_s) = \delta(\bm{W}_s \bm{M} \bm{D})$ where, $v$ is the size of the word vocabulary and $\delta$ is a $\tanh$ activation. In our experiments, we find a trainable $g(\bm{W}_s)$ achieves better performance.

We propose a two step training procedure. The first step involves training the base model, where Eq. \ref{eq:pred_seen} is trained with only images and semantics of seen classes. In the second step of fine-tuning, when novel class information becomes available, we replace $\bm{W}_s$ by $\bm{W}$ and train it with few-shot examples. Eq.~\ref{eq:pred_seen} then becomes $\bm{p} = \sigma \big(f(\bm{X})^T \bm{U}  g(\bm{W}) \big)$, where, $\bm{p}$ contains scores of both seen and novel classes. In this manner, model is quickly adapted for novel classes. 

Notably, 
although our network can predict scores for all classes, no new tunable weights are added.
In both steps, our network tr{ies} to align the feature 
with its corresponding semantics. From the network's perspective it does not matter how many classes are present. It only cares how compatible a feature is with the corresponding class semantics. This is why our network does not forget the seen classes as these semantic prototypes serve as an anchor to retain previous knowledge. Adding new classes is therefore not a new task for the network. During fine-tuning, the model still performs the same old task of aligning feature and semantics.

\subsubsection{Learning with Imbalance Data}
After base training on seen classes, novel classes information becomes available during inference stage i.e., few-shot images and the word vectors of both zero and few-shot classes. Here, the few-shot image may contain seen instances as well. In this way, the fine-tuning stage contains an imbalanced data distribution and the model must minimize bias towards the already seen classes. To this end, we propose a rebalancing loss  for the fine-tuning stage.

Suppose, $p \in \bm{p}$ is the alignment score for a visual feature and the corresponding  class semantics. As the fine-tuning is done on rare data, we need to penalize the cross-entropy (CE) loss based on the quality of alignment. If the network makes a mistake, we  increase the penalty and if the network is already performing well, we employ a low or negative penalty. Suppose, the penalty $h(\cdot)$ is a function of $p$ and $p_*$ where $p_*$ determines the penalization level, then the loss is,
\begin{equation}
    L(p) = - \log p + \beta\, h(p,p_*), \label{eq:initial_loss}
\end{equation}
where, $\beta$ is a hyper-parameter. Here, $h(p,p_*)$ is given by, 
\begin{equation}
    h(p,p_*) = \log (1 + p_* - p), \label{eq:penalty}
\end{equation}
where $(p_* {-} p)$ represents the violation of the expected alignment that controls the margin of the penalty function. We explore two alternatives for selecting $p_*$, a fixed value in range $0 {<} p_* {\leq} 1$ and a dynamically adjusted value based on $p_* = \max_{i \in C} p_i$. We can get the following scenarios based on the choice of $p_*$ and positive alignment quality $p$:\vspace{-0.5em}
\begin{itemize}
    \item  \emph{Expected case, $p {>} p_*$:} Negative penalty to calculate lower loss compared to CE (lower bounded by 0).
\item \emph{Moderate case, $p {=} p_*$:} Zero penalty and the calculated loss is equal to regular CE.
\item \emph{Extreme case, $p {<} p_*$:} High penalty in comparison to regular CE loss. 
\end{itemize}
Plugging the penalty definition from 
Eq.~\ref{eq:penalty} to Eq.~\ref{eq:initial_loss} and enforcing positive loss values $L(p) \in \mathbb{R}^{+}$, we obtain, 
\begin{align}
 L(p) = \max \Bigg[0,- \log \frac{p}{(1 + p_* - p)^\beta}\Bigg].
\end{align}
After adding  {an} $\alpha$-balanced modulating factor from focal loss \cite{RetinaNet_2017_ICCV}, we have,
\begin{align}\label{eq:new_loss}
  {L(p)} = \max \Big[ 0,-\alpha_t (1-p_t)^{\gamma} \log p_t \Big], \\ 
  \text{where } p_t =\begin{cases}
    								 \frac{p}{(1 + p_* - p)^\beta}, & \text{if $y=1$}\\ \notag
    								1-p, & \text{otherwise}.
 \end{cases}
\end{align}
Here, $\beta$ is the focusing parameter that focuses on hard cases and $y$ is the corresponding value from the one-hot encoded ground-truth vector.
With $\beta{=}0$, Eq.~\ref{eq:new_loss} becomes equivalent to focal loss and with $\beta{=}0$, $\gamma{=}0$, $\alpha{=}1$, Eq.~\ref{eq:new_loss} becomes 
%NBthe
cross-entropy loss.

Since several objects can co-occur in the same scene, the fine-tuning data can have seen instances. To emphasis{e} rare classes more than the seen ones, we apply our rebalancing loss only on the novel class examples. For a seen anchor, only the focal loss is calculated. Thus, the final loss is,
\begin{align}
    L = \lambda L(s) + (1 - \lambda) L(n).
\end{align}
For the case of $L(s)$ and $L(n)$,  $\beta {=} 0$ and $\beta {>} 0$ respectively. $L(s)$ and $L(n)$ represent the compatibility scores of seen and novel (few-shot and unseen) classes i.e., $s \in \{1,2, ..,S\}$ and $n \in \{1,2,..,T\}$.

During inference when a test image is presented, a simple forward pass provides compatibility scores of seen, few-shot and unseen classes for each bounding box. If the score is higher than a threshold, we consider it a correct detection.

\textbf{Analysis:} Based on the quality of alignment, our proposed loss penalizes positive anchors. This scenario helps in  {the} class imbalance problem. Especially, in the extreme case, when the network fails to detect a positive few-shot anchor, we highly penalize our network predictions. It gives
extra supervision to the network that it must not make errors on the few-shot classes. In contrast, for the expected and moderate cases, we reduce the loss which avoids the network becoming too
confident on few-shot examples. Since, unseen objects are more related to the seen objects as compared to background, a low penalty on confident cases implicitly promotes discovering unseen classes. In effect, this leads to low overfitting on the few-shot classes that helps in achieving good performance on totally unseen classes.

\textbf{Loss shape analysis:} In Fig.~\ref{fig:loss_shape}, we visualize the loss. 
Fig.~\ref{fig:loss_shape} (a) shows how the shape of binary cross entropy (CE) changes
with different values of $\beta$. $\beta$ controls the penalty $h(.)$ which modifies CE loss. For a fixed $p_*{=}1$, increasing $\beta$ calculates a higher penalty for any mistake in the prediction. For a fixed margin penalty $p_*{=}0.5$ in Fig.~\ref{fig:loss_shape} (b), a network can predict a lower, equal and higher score than $p_*$. Correspondingly, it enables the network to calculate a less, equal and higher penalty for expected, moderate and extreme cases, respectively. In contrast, for the dynamic margin penalty case Fig. \ref{fig:loss_shape} (c), the predicted score can be at best $p_* = \max_{i \in C} p_i$. Therefore, the extreme case works similarly to the fixed $p_*$ scenario but for the other cases, the network calculates a loss equal to CE/Focal loss. The dynamic $p_*$ estimates the quality of predicted scores based on the current anchor specific situation. For example, for a given anchor, a small predicted score  (e.g., $0.1$) for the ground-truth class can be considered as good prediction if all other predictions are $<0.1$. It helps to make a good balance between seen, few-shot and unseen predictions because the loss does not unnecessarily tries to maximize the ground-truth class score and thus avoids over-fitting.

\subsection{Implementation Details}
We implement our framework with a modified version of the RetinaNet architecture proposed in \cite{rahman2018polarity} (see Fig.~\ref{fig:arch}(a)). It incorporates the word vectors at the penultimate layers of classification and regression subnets. While performing the base training with focal loss at the first step, we follow the recommended process in  \cite{rahman2018polarity}, where only seen word vectors are used in word processing network, $g(.)$ (see Fig.~\ref{fig:arch}(b)). During the fine-tuning step, we update the base model with newly available data and our proposed loss function. As shown in Fig.~\ref{fig:arch}(c), fine-tuning uses both seen and novel word vectors inside $g(.)$. Note that, in addition to novel class data, the small dataset used for fine-tuning includes some seen instances as well. We train our model for 10 epochs during the fine-tuning stage. 
After the fine-tuning is finished, our framework can detect seen, few-shot, and unseen classes simultaneously. We use the Adam optimizer for each training stage.

\begin{table*}[!t]\setlength{\tabcolsep}{5pt}
\begin{minipage}{0.48\textwidth}
\caption{\small ASD performance.}
\label{tab:asd_result}
% \vspace{-0.8em}
\centering
\scalebox{0.65}{
\begin{tabular}{|c|c||c|c|c|c|c|c|c|}
\hline
    \#-&\multirow{2}{*}{Method}&   \multicolumn{3}{|c|}{ASD} & \multicolumn{4}{|c|}{GASD}\\  \cline{3-9}
      Shot&  	&  unseen & FS & HM	& seen & unseen&FS & HM\\ \hline
      \multirow{3}{*}{1}&Baseline-I&3.74&1.60&2.25&\textbf{54.11}&2.04&0.73&1.60 \\
      &Baseline-II&8.57&21.39&12.23&51.89&3.79&9.62&7.74 \\      
      &Ours&\textbf{16.57}&\textbf{23.50}&\textbf{19.44}&51.70&\textbf{10.75}&\textbf{11.83}&\textbf{15.23} \\ \hline
      \multirow{3}{*}{5}&Baseline-I&4.16&2.69&3.27&\textbf{54.15}&2.35&1.20&2.35 \\
      &Baseline-II&8.69&26.19&13.05&51.67&4.85&18.20&10.70 \\     
      &Ours&\textbf{18.22}&\textbf{26.31}&\textbf{21.53}&51.18&\textbf{12.70}&\textbf{18.34}&\textbf{19.63} \\ \hline
      \multirow{3}{*}{10}&Baseline-I&3.45&2.95&3.18&\textbf{54.25}&1.89&1.56&2.53 \\
      &Baseline-II&7.26&31.14&11.78&51.00&4.12&25.00&9.91 \\    
      &Ours&\textbf{13.21}&\textbf{33.52}&\textbf{18.95}&51.18&\textbf{9.71}&\textbf{26.96}&\textbf{18.79} \\ \hline
% \hline
\end{tabular}}\vspace{-0.8em}
\end{minipage}
%\hfill
\hspace{1em}
\begin{minipage}{0.48\textwidth}\setlength{\tabcolsep}{5pt}
\caption{\small Ablation study with 5-shot case.}
\label{tab:ablation}
%\vspace{-0.8em}

\scalebox{0.65}{
\begin{tabular}{|c||c|c|c|c|c|c|c|}
\hline
      \multirow{2}{*}{Method}&   \multicolumn{3}{|c|}{ASD} & \multicolumn{4}{|c|}{GASD}\\  \cline{2-8}
      	&  unseen & FS & HM	& seen & unseen&FS & HM\\ \hline
      Baseline-I&4.16&2.69&3.27&\textbf{54.15}&2.35&1.20&2.35 \\
      Baseline-II&8.69&26.19&13.05&51.67&4.85&18.20&10.70 \\ \hline
      Ours with FL&13.69&23.81&16.61&51.20&9.21&16.34&15.85 \\ %\cline{2-8}
      Ours with AL&7.03&24.17&10.89&50.74&5.94 &17.46&12.23\\ \hline
      Ours ($p_*=0.3$)&17.20&26.85&20.97&51.48&11.84&19.21&19.24 \\ 
      Ours ($p_*=0.5$)&15.24&24.02&18.65&50.65&10.38&17.06&17.17 \\ %\cline{2-8}
      Ours ($p_*=1.0$)&16.17&27.17&20.27&50.58&11.29&19.83&18.90 \\ \hline %\cline{2-8}
      Ours*&16.60&24.05&19.64&51.32&11.09&16.71&17.70\\ %\hline
      Ours&\textbf{18.22}&\textbf{26.31}&\textbf{21.53}&51.18&\textbf{12.70}&\textbf{18.34}&\textbf{19.63}\\ \hline
      % Our&&&&&&&\\ \hline
\end{tabular}}\vspace{-0.8em}
\end{minipage}
\end{table*}

\begin{figure*}[!tp]
    \centering
    \includegraphics[width=1\textwidth,trim={4.2cm 4.8cm 3.2cm 4.9cm},clip]{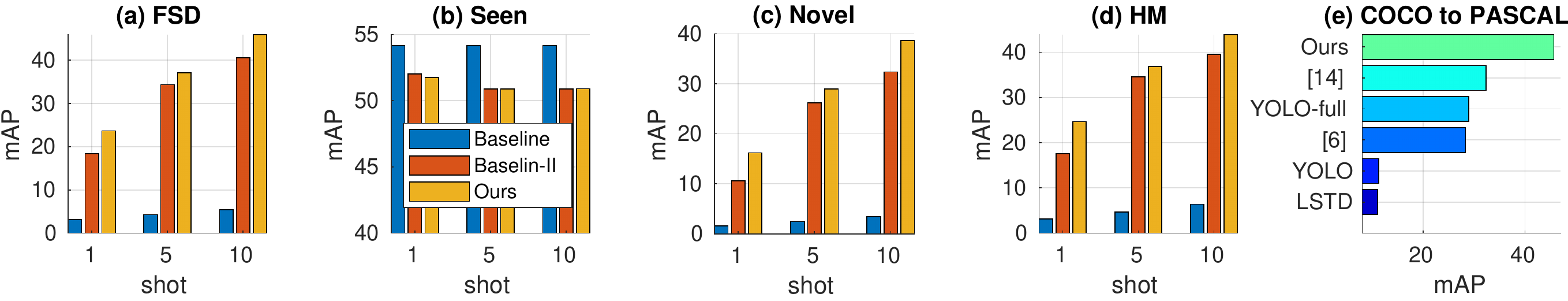}\vspace{-1em}
    % left lower right up
    \caption{\small FSD performance. (a) 1-, 5- and 10-shot detection mAP, (b), (c) and (d) seen, unseen and harmonic mean (HM) of GFSD. (e) 10-shot detection mAP with MSCOCO to PASCAL}
    \label{fig:fsd}
\end{figure*}

\section{Experiments}

\textbf{Datasets:} We evaluate our work on the MSCOCO-2014 \cite{MSCOCO_2014} and PASCAL VOC 2007/12 datasets. For the MSCOCO experiment, we adopt the 65/15 seen/unseen split setting used in \cite{rahman2018polarity,Rahman_2019_ICCV}. In both ZSD and FSD experiments, we consider unseen classes as the novel ones. However, in ASD experiments, we further split 15 novel classes into 8 few-shot and 7 unseen classes. We use the same \num{62300} images during training where no single instance of novel classes is present. For testing ASD, we use \num{10098} images where each image contains at least one novel object. However for GASD testing, we use the entire validation set of \num{40504} images.
For the PASCAL VOC 2007/12 experiment, we adopt 15/5 seen/novel split settings from \cite{Kang_2019_ICCV}. As recommended, we use train$+$val sets from PASCAL VOC 2007 and 2012 as training data and test-set from PASCAL VOC 2007 for evaluation. For FSD/ASD, we randomly choose additional images with a few (1/5/10) annotated bounding boxes for each novel category. These images may contain some seen objects as well.  
For both MSCOCO and PASCAL VOC classes and vocabulary texts,  
we use 300-dimensional and $\ell_2$ normalized word2vec vectors \cite{Mikolov_NIPS_2013}. We have used same set of \num{4717} vocabulary atoms as used in \cite{rahman2018polarity} which are originally taken from \cite{NUS_WIDE_09}.

\textbf{Evaluation criteria:} {For FSD and ASD, we evaluate our method with mean average precision (mAP). To report GFSD and GASD, we calculate the harmonic mean of the individual seen and novel class mAP. For ZSD, we report recall@100 results as recommended in \cite{Bansal_2018_ECCV}.}

\textbf{Validation experiment:} $\alpha, \beta, \gamma$ and $\lambda$ are the hyper-parameter of our model. Among them, $\alpha$ and $\gamma$ are spefici to focal loss. %hyper-parameters. 
Thus, we fix the recommend value $\alpha = 0.25$ and $\gamma = 2$ following % original RetinaNet model 
\cite{RetinaNet_2017_ICCV}. {We validate $\beta$ and $\lambda$ by creating a validation dataset based on splitting 65 seen classes into 55 seen and 10 novel classes.  From {the} validation experiment, we select $\beta=5$ and $\lambda=0.1$. Detailed validation results are presented in the appendix.}

\textbf{Baseline methods:} Here, we introduce the baseline methods used for comparison in this work. \\
$\bullet$ \emph{Baseline-I:} A RetinaNet architecture where fixed semantics are used in semantic processing pipeline i.e. $g(\bm{W}_s) = \bm{W}_s$ and the training is done with the basic focal loss. The fine-tuning step uses all seen and novel class data together.\\
$\bullet$ \emph{Baseline-II:} The second baseline approach is identical to Baseline-I, except that the fine-tuning step uses novel class data and a few examples of seen.\\
Finally, \emph{Ours} denote the complete approach where the RetinaNet architecture is trained with adaptive  prototypes in the semantic processing pipeline i.e.  $g(\bm{W}_s) = \delta(\bm{W}_s \bm{M} \bm{D})$ and the training is done with our proposed loss.

\begin{table*}[!t]
  \caption{\small FSD mAP on Pascal VOC 2007 test set.
  } 
  \label{tab:pascal_perclss}
  \vspace{-1.1em}
  \begin{center}\setlength\tabcolsep{5pt}
  \scalebox{.7}{
    \begin{tabular}{|c|c|ccccccccccccccc|c||ccccc|c|}
    \hline
    
\multicolumn{2}{|c|}{}  & \multicolumn{16}{|c||}{Base} & \multicolumn{6}{|c|}{Novel} \\ \hline     
    
\rotatebox{90}{\# shot}&\rotatebox{0}{\makecell{ \textbf{Method}}}&\rotatebox{90}{aero}&\rotatebox{90}{bike}&\rotatebox{90}{boat}&\rotatebox{90}{bottle}&\rotatebox{90}{car}&\rotatebox{90}{cat}&\rotatebox{90}{chair}&\rotatebox{90}{table}&\rotatebox{90}{dog}&\rotatebox{90}{horse}&\rotatebox{90}{person}&\rotatebox{90}{plant}&\rotatebox{90}{sheep}&\rotatebox{90}{train}&\rotatebox{90}{tv} &\rotatebox{90}{mean}&\rotatebox{90}{bird}&\rotatebox{90}{bus}&\rotatebox{90}{cow}&\rotatebox{90}{mbike}&\rotatebox{90}{sofa}&\rotatebox{90}{mean} \\  \hline
\multirow{3}{*}{3}&LSTD \cite{chen2018lstd} &74.8 &68.7 &\textbf{57.1} &44.1 &\textbf{78.0} &83.4 &46.9 &64.0 &\textbf{78.7} &\textbf{79.1} &70.1 &39.2 &58.1 &\textbf{79.8} &71.9 &66.3& 23.1 &\textbf{22.6} &15.9 &0.4 &0.0 &12.4\\
%&&&&&&&&&&&&&&&&&&&&&&\\
&Kang \etal \cite{Kang_2019_ICCV}&73.6&\textbf{73.1}&56.7&41.6&76.1&78.7&42.6&\textbf{66.8}&72.0&77.7&68.5&42.0&57.1&74.7&70.7&64.8&26.1&19.1&\textbf{40.7}&20.4&\textbf{27.1}&26.7\\
 &Ours& \textbf{80.4}&52.8&50.2& \textbf{55.9}&76.9& \textbf{85.1}&\textbf{49.8}&54.0&76.8&72.7&\textbf{81.1}&\textbf{44.8}&\textbf{61.7}&79.0& \textbf{76.8}&\textbf{66.5}& \textbf{28.6}&18.7&15.5& \textbf{55.5}&17.4& \textbf{27.1}\\ \hline
\multirow{3}{*}{5}&LSTD \cite{chen2018lstd} &70.9 &71.3 &\textbf{59.8} &41.1 &\textbf{77.1} &81.9 &45.1 &\textbf{67.2} &78.0 &\textbf{78.9} &70.7 &41.6 &63.8 &\textbf{79.7} &66.8 &66.3& 22.8 &52.5 &31.3 &45.6 &\textbf{40.3} &38.5\\
%&&&&&&&&&&&&&&&&&&&&&&\\ \hline
&Kang \etal \cite{Kang_2019_ICCV}& 65.3& \textbf{73.5}& 54.7& 39.5& 75.7& 81.1& 35.3& 62.5& 72.8& 78.8& 68.6& 41.5& 59.2& 76.2& 69.2& 63.6&30.0&\textbf{62.7}& \textbf{43.2}& 60.6& 39.6& 47.2 \\
&Ours&\textbf{78.8}&50.2&56.5&\textbf{59.7}&75.8&\textbf{85.4}&\textbf{52.1}&54.0&\textbf{78.9}&72.4&\textbf{81.8}&\textbf{47.1}&\textbf{64.1}&73.5&\textbf{76.6}&\textbf{67.1}&\textbf{47.3}&50.3&39.6&\textbf{68.9}&39.9&\textbf{49.2}\\ \hline

    \end{tabular}}
  \end{center}
  % \vspace{-1.5em}
\end{table*}

\subsection{ASD Performance} 
Here, we discuss the ASD and GASD performance with the 65/8/7 split of MSCOCO. For ASD, we show the performance of novel classes (i.e. unseen and few-shot classes) and the harmonic mean of individual performances. For GASD, we report separate seen, few-shot, unseen mAP and their harmonic mean mAP to show the overall performance.

\begin{figure*}[!t]
    \centering
    \includegraphics[width=1\textwidth]{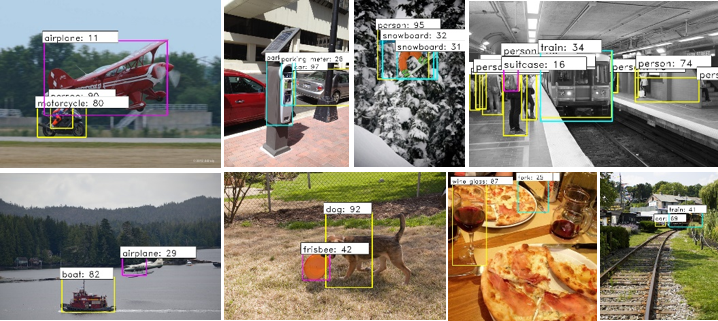}
    \vspace{-2em}
    \caption{\small Qualitative results for Generalized ASD. Yellow, blue and pink  boxes represent seen, few-shot and unseen objects, respectively. \emph{(best viewed with zoom)}
    }
    \label{fig:qualitative}
\end{figure*}

\begin{figure*}[!t]\vspace{-0.5em}
    \centering
    \includegraphics[width=1\textwidth, clip=true, trim = 2mm 2mm 2mm 2mm]{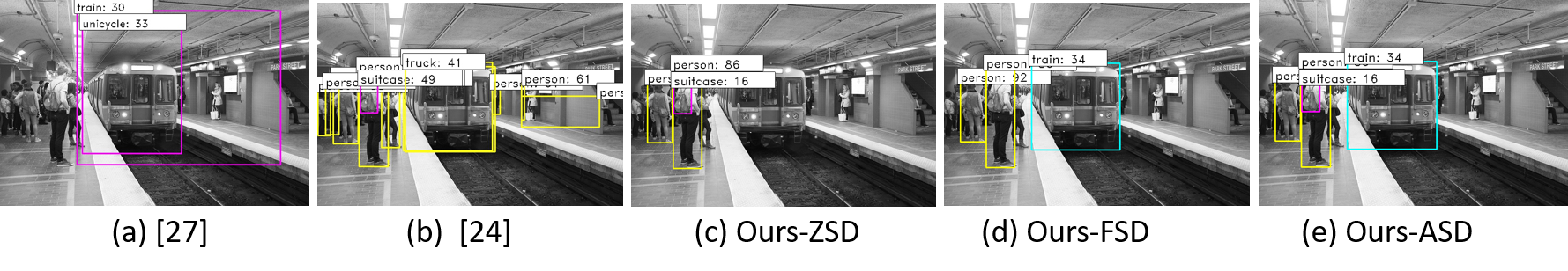}\vspace{-1.0em}
    \caption{\small Qualitative comparison with \cite{rahman2018ZSD}, \cite{rahman2018polarity} and Our method. Object bounding boxes: Yellow (seen),
blue (few-shot) and pink (unseen). \emph{(best viewed with zoom)}}
    \label{fig:qualitative_comparison}
\end{figure*}

\textbf{Main results:} In Table \ref{tab:asd_result}, we report the our main results and comparisons with the baselines. Our observations are as follows: \textbf{(1)} Using more few-shot examples generally helps.
However the effect of higher shots on the unseen performance is not always positive since more instances of few-shot classes can bias the model towards them.
\textbf{(2)} Except Baseline-1, few-shot mAP is always better than unseen mAP because few-shot  examples with our proposed loss improve the alignment with respective class semantics. In {the} Baseline-I case, as all seen and few-shot data is used together, the network overfits to seen classes. \textbf{(3)} Our seen class performance in GASD remains good across different shots. This denotes  that the network does not forget seen classes when trained on novel ones. Seen classes get the maximum performance for the Baseline-I due to overfitting, thereby giving %NB a 
poor performance on novel classes. \textbf{(4)} Across different shots, Baseline-II beats {the} Baseline-I method as it is less prone to overfitting. With the proposed adaptive semantic prototypes and our loss function, we beat Baseline-II. \textbf{(5)} The improvement for unseen mAP is greater than  few-shot or seen mAP irrespective of the number of shots, ASD, or GASD tasks. It tells us that our loss formulation not only tackles the class imbalance of few-shot classes but also promotes detection of unseen classes. In Fig.~\ref{fig:qualitative} and ~\ref{fig:qualitative_comparison}, we show qualitative results for GASD.

\textbf{Ablation studies:} In Table \ref{tab:ablation}, we report ablation experiments with with alternate versions of our approach. Baseline-I and Baseline-II operate with fixed semantics. 
For the rest of the cases, we use our adaptive semantic-prototype approach (Sec.~\ref{sec:dontforget}) to update the word vectors. Here, we first use a basic focal loss \cite{RetinaNet_2017_ICCV} (FL) to train the network. This approach outperforms both baselines because of the adaptable semantic prototypes. Then, we try two variants of FL: Anchor Loss \cite{Ryou_2019_ICCV} (AL) and a modified anchor loss with our loss penalty definition for few-shot classes. We notice that these variations do not work well in both ASD and GASD cases because AL penalizes negative anchors that network confuses with positive ones. This idea is beneficial for traditional recognition cases, but unsuitable for ZSD/FSD/ASD scenarios. This is because a negative anchor may contain an unseen object which is closely related to seen or few-shot semantics, and we do not want to suppress the anchor even though it predicts similar scores as the positive ones. Next, we apply our loss by fixing $p_*$ to a constant value e.g., 0.3, 0.5, and 1. These trials outperform both baselines and FL based methods since the network emphasizes 
few-shot examples based on the quality of the visual-semantic alignment. Finally, alongside the adaptive semantics, we apply our final loss formulation which dynamically selects $p_*$. Our loss beats all previously described settings because it brings better emphasis on novel classes. Notably, we also experiment with the Our* case that applies our loss to all predictions (instead of just the novel scores) i.e., $\beta >0$ for all classes. However, it does not perform as well
as Ours 
potentially because the representations suitable for seen classes are already learnt well.

\subsection{FSD Performance}

If 
$\uC{=}0$, our proposed ASD framework becomes a few-shot detection problem. In this paper, we experiment on FSD with the following three different dataset setups.

\textbf{MSCOCO:} Here we consider all 15 novel classes of the MSCOCO split \cite{rahman2018polarity} as few-shot classes. We report mAP results of 1, 5 and 10-shot detection tasks of 
Baseline-I, Baseline-II, and Ours model in Fig.~\ref{fig:fsd} (a-d). Besides, we report generalized FSD results. Overall, FSD performance 
improves with more examples. When trained with the adaptive semantic prototypes and rebalancing loss, our model successfully outperforms both baselines.

\textbf{MSCOCO to PASCAL:} It is a cross-dataset experiment. Following  \cite{Kang_2019_ICCV}, we use 20 PASCAL VOC classes (overlapped with MSCOCO) as the few-shot classes and the 
% NB rest
{remaining} 60 MSCOCO classes as seen. This setting performs  base training on MSCOCO seen classes and fine-tunes the base model using {the} 10-shot examples of the PASCAL VOC dataset. Finally, the PASCAL VOC 2007 test set is used to evaluate 
%NB the 
FSD. In Fig.~\ref{fig:fsd}(e), our method outperforms others including a recent approach \cite{Kang_2019_ICCV}.

\textbf{PASCAL VOC 2007/12:} Using the 15/5 novel-split-1 proposed in \cite{Kang_2019_ICCV}, we compare FSD performance of our work with Kang \etal \cite{Kang_2019_ICCV} and LSTD \cite{chen2018lstd} in Table \ref{tab:pascal_perclss}. We achieve better performance than them in both novel and base class detection with 3- and 10-shot detection settings.

\begin{table}[!t]\setlength{\tabcolsep}{5pt}
\centering
\caption{\small Recall scores for ZSD and GZSD tasks on MSCOCO dataset.}
\label{tab:zsdmap}
%\vspace{-0.8em}
\scalebox{0.9}{
\begin{tabular}{|c|c|c|c|c|}
\hline
\multirow{2}{*}{\textbf{Method}}   & \multirow{2}{*}{\textbf{ZSD}} & \multicolumn{3}{c|}{\textbf{GZSD}}\\ \cline{3-5}
 &   & Seen	& Unseen & HM \\ \hline

SB~\cite{Bansal_2018_ECCV} &24.39&-&-&- \\ %\hline
DSES~\cite{Bansal_2018_ECCV} &27.19&15.02&15.32&15.17 \\ %\hline
SAN~\cite{rahman2018ZSD}&12.27&-&-&-\\ \hline
% ZSD-Textual~\cite{Li_AAAI_2019}&34.3&-&-&-\\ %\hline
Baseline	&18.67&\textbf{42.21}&17.60&24.84 \\ %\hline
Ours &\textbf{32.83}&41.66&\textbf{27.34}&\textbf{33.01}\\ \hline

\end{tabular}}
% \vspace{-0.5em}
\end{table}

\subsection{ZSD Performance}
For ZSD case, after the first step of base training, we do not have any more data to fine-tune. Thus, we perform the second step of fine-tuning with the same data used in the first step but apply our loss instead of the focal loss. As $\fC{=}0$, we consider each seen class as a few-shot class during the second step. It emphasizes all seen classes in the same way. But, based on the dynamic choice of $p_*$, the network penalizes a bad prediction by calculating high loss and compensates a good prediction with no penalty. % that works the same as focal loss. 
We notice that it helps to improve ZSD performance. We apply this process with the 
%65/15 split setting of \cite{rahman2018polarity} and 
48/17 split setting of  
%NB of 
\cite{Bansal_2018_ECCV} on MSCOCO. We report 
the result of this experiment
in Table \ref{tab:zsdmap}. As recommended in %\cite{rahman2018polarity} and 
\cite{Bansal_2018_ECCV}, we compare our work using recall measure where it outperforms other methods. 

\section{Conclusion}
In this paper, we propose a unified any-shot detection approach where novel classes include both unseen and few-shot objects. Traditional approaches consider solving zero and few-shot tasks separately, whereas our approach encapsulates both tasks into a common framework. This approach does not forget the base training while learning novel classes, which helps to perform generalized ASD. Moreover, we propose a new loss function to learn new tasks. This loss penalizes the wrong prediction of a novel class more than the seen classes. % and vice versa. 
We evaluate the proposed ASD tasks on the challenging MSCOCO and PASCAL VOC datasets. Besides, we compare ZSD and FSD performance of our approach with established state-of-the-art methods. Our first ASD framework delivers strong performance in ZSD, FSD, and ASD tasks compared to previous methods.

{\small
\bibliographystyle{ieee_fullname}
\bibliography{bib_file}
}

%-------------------
\newpage

\section{Appendix A}

% \author{Shafin Rahman$^{*\dagger}$, Salman Khan$^{\ddagger}$, Nick Barnes$^{*}$, Fahad Shahbaz Khan$^{\ddagger}$\\
% $^{*}$Australian National University, $^{\dagger}$North South University, \\
% $^{\ddagger}$Inception Institute of Artificial Intelligence\\
% {\tt\small {\{shafin.rahman,nick.barnes\}@anu.edu.au, \{salman.khan,fahad.khan\}@inceptioniai.org}}
% }
% \maketitle

This appendix material provides additional qualitative results, gradient analysis for the proposed loss function and validation experiments for the hyper-parameters. 

\begin{figure*}[h]
    \centering
    \includegraphics[width=1\textwidth,trim={0cm 0cm 5.1cm 0cm},clip]{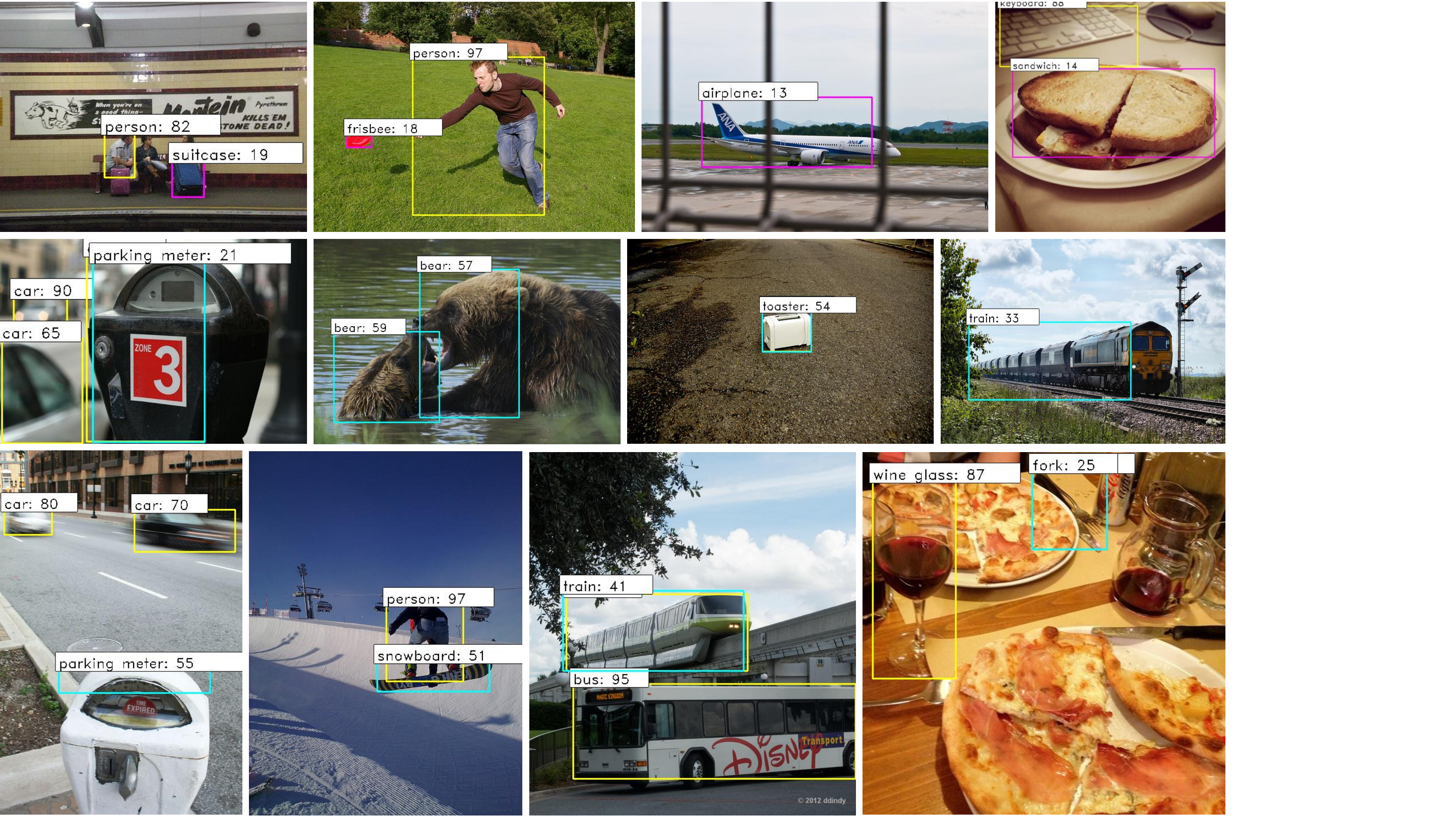}
    % left lower right up
    % \vspace{-0.5em}
    \caption{More qualitative results for generalized ASD. Yellow, blue and pink  boxes represent seen, few-shot and unseen objects, respectively. The proposed approach can detect classes from all the three categories. 
    }
    \label{fig:qualitative2}
\end{figure*}

\subsection{Qualitative Results}
In Fig.~\ref{fig:qualitative2}, we show additional qualitative results of our approach on the ASD task. All visual results are generated with a single ASD model that can detect seen, unseen and few-shot classes, simultaneously.

\subsection{Gradient analysis} In this section, we derive the gradients of our proposed loss w.r.t $p$. For simplicity, we only focus on the predictions corresponding to the ground-truth class. We show the gradient curve in Fig. 4(d) of the main paper. 
\begin{align}%\scriptsize
 \frac{\partial L}{\partial p} = 
\begin{cases}
    \dfrac{1}{p\left(p-p_*-1\right)\left(p-\left(1-p+p_*\right)^{\beta}\right)} \Big[ \alpha_t\big(\left(\beta-1\right)p \\
    \qquad\qquad + p_*+1\big)\left(1- \frac{p}{\left(1-p+p_*\right)^{\beta}}\right)^{\gamma}\big(\gamma p \log \left(p \right) \\
    \qquad\qquad +\left(1-\beta {\gamma} \log \left(1-p+p_*\right)\right)p-\left(1-p+p_*\right)^{\beta}\big) \Big] \\
    \qquad\quad \text{if }\;  \big(\alpha_t \beta\left(1-\frac{p}{(1-p+p_*)^{\beta}}\right)^{\gamma} \log \left(1-p+p_* \right) \\
    \qquad\qquad  - \alpha_t \left(1-\frac{p}{(1-p+p_*)^{\beta}} \right)^{\gamma} \log(p)\big) > 0 \text{ and } y=1\\
    \dfrac{\alpha_t p^{\gamma}}{1-p}- \alpha_t \gamma \log \left(1-p \right)p^{\gamma-1} \\
    \qquad\quad \text{if }\; \big(\alpha_t p^{\gamma} \log(1-p)\big) < 0 \text{ and } y=0\\
    0  \qquad\;\, \text{otherwise}.
\end{cases}  \notag
\end{align}

\subsection{Validation Set Experiments} 
Note that $\alpha, \beta, \gamma$ and $\lambda$ are the hyper-parameter of our model. Among them, $\alpha$ and $\gamma$ are focal loss hyper-parameter. Thus, we fix the recommended value of $\alpha = 0.25$ and $\gamma = 2$ as per the original RetinaNet model \cite{RetinaNet_2017_ICCV}. $\beta$ and $\lambda$ are the hyper-parameters proposed in our loss formulation. Therefore, we validate $\beta$ and $\lambda$ by further splitting 65 seen classes into 55 seen and 10 novel classes to create a validation set. Then, we perform a 5-shot detection task on this validation split. We report the validation results in Table \ref{tab:validation}. From our validation study, we select $\beta=5$ and $\lambda=0.1$ and use this value in all of our experiments.

\begin{table*} [h]
\begin{center}
\begin{tabular}{|l|c|c|c|c|c|c|c|c|}
\hline
      \backslashbox{$\beta$}{$\lambda$}&0.1&0.2&0.3&0.4&0.5&0.7&0.9&1.0 \\ \hline
    0.5&30.90&30.22&28.70&26.92&24.84&22.53&17.23&0.23 \\ \hline
      1&29.70&28.82&29.25&28.06&26.96&23.73&16.83&0.24 \\ \hline
      2&30.04&29.78&29.07&29.93&27.05&23.99&18.42&0.13 \\ \hline
      5&\textbf{31.54}&31.54&30.50&29.52&27.64&25.83&17.14&0.26 \\ \hline
\end{tabular}
  \end{center}
\caption{Validation studies with 5-shot detection performance}
\label{tab:validation}
\end{table*}

\end{document}